\pdfoutput=1

\documentclass[11pt]{article}

\usepackage{acl}

\usepackage{times}
\usepackage{latexsym}
\usepackage{caption}
\usepackage{graphicx}
\usepackage{subcaption}
\usepackage{csquotes}
\usepackage{amsmath, mathtools, multirow}
\usepackage{url}
\usepackage{tabularx}

\usepackage[T1]{fontenc}

\usepackage[utf8]{inputenc}

\usepackage{microtype}

\newcommand{\sara}[1]{{\textcolor{black}{#1}}}
\newcommand{\mg}[1]{{\textcolor{black}{#1}}}

\title{Efficient yet Competitive Speech Translation: FBK@IWSLT2022}

\author{Marco Gaido\textsuperscript{1,2 \includegraphics[scale=0.04]{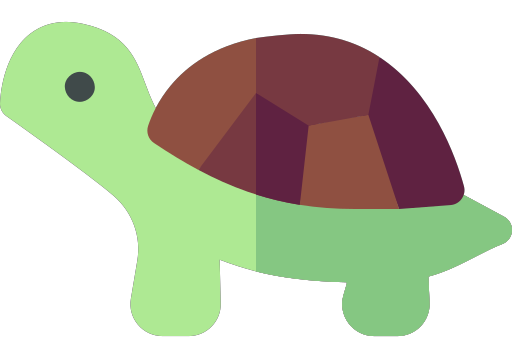}}, Sara Papi\textsuperscript{1,2 \includegraphics[scale=0.04]{imgs/turtle.png}}, Dennis Fucci\textsuperscript{1,2}, Giuseppe Fiameni\textsuperscript{3}, \\
{\bf Matteo Negri\textsuperscript{1}}, {\bf Marco Turchi\textsuperscript{1}} \\
  \textsuperscript{1}Fondazione Bruno Kessler\\
  \textsuperscript{2}University of Trento\\
  \textsuperscript{3}NVIDIA AI Technology Center\\
  \texttt{\{mgaido, spapi, dfucci, negri, turchi\}@fbk.eu} \\
  }

\begin{document}
\maketitle
\begin{abstract}
\newcommand\blfootnote[1]{%
  \begingroup
  \renewcommand\thefootnote{}\footnote{#1}%
  \addtocounter{footnote}{-1}%
  \endgroup
}
\blfootnote{\includegraphics[scale=0.04]{imgs/turtle.png} The authors contributed equally.}
The primary goal of this FBK's systems submission to the IWSLT 2022 offline and simultaneous speech translation tasks is to reduce model training costs without sacrificing translation quality. As such, we first question the need of ASR pre-training, showing that it is not essential to achieve competitive results. Second, we focus on data filtering, showing that a simple method that looks at the ratio between source and target characters yields a quality improvement of 1 BLEU. Third, we compare different methods to reduce the detrimental effect of the audio segmentation mismatch between training data manually segmented at sentence level and inference data that is automatically segmented. Towards the same goal of training cost reduction, we participate in the simultaneous task with the same model trained for offline ST. The effectiveness of our lightweight training strategy is shown by the high score obtained on the MuST-C en-de corpus (26.7 BLEU) and is confirmed in high-resource data conditions by a 1.6 BLEU improvement on the IWSLT2020 test set over last year's winning system.

\end{abstract}

\section{Introduction}

The yearly IWSLT offline speech translation (ST) evaluation campaign aims at comparing the models produced by companies, universities, and research institutions on the task of automatically translating 
speech
in one language into 
text
in another language.
Given a blind test set, participants' submissions are ranked according to the obtained SacreBLEU score \cite{post-2018-call}.

Over the years, the competition to achieve the highest score has driven to bigger and bigger models trained on large datasets: the 2021 winning model \cite{bahar-etal-2021-without} has twice the number of encoder layers (12 vs 6), and a deeper (6 vs 4 layers) and larger (1024 vs 512 features) decoder compared to the 2019 winner \cite{potapczyk-etal-2019-samsungs}.
In addition, most of the competitors have relied on knowledge transfer techniques \cite{ansari-etal-2020-findings,iwslt_2021}, such as the initialization of the ST model encoder with the encoder of an ASR system trained on large corpora \cite{bansal-etal-2019-pre}.
All these practices have contributed to a remarkable increase in computational expenses and energy consumption that are antithetic with the recent
rise of concerns on the social and environmental consequences of these costs
\cite{strubell-etal-2019-energy}.

Among the
harms  
inherent to the
high computational cost of training ST systems, there is also the risk of 
restricting the participation in
competitions like IWSLT to 
few big players 
from the industry sectors that can afford them.
As part of a research institution, with this work we try to answer the question: \textit{can we reduce the training cost of ST systems without 
sacrificing final translation quality?} Specifically, can we train a competitive direct ST model from scratch, without expensive pre-training (e.g. ASR pre-training or self-supervised learning on huge dataset -- \citealt{BaevskiZMA20-wav2vec})?


To answer these questions, we perform a preliminary study on the English-German (en-de) section of MuST-C \cite{Cattoni2020mustc-v2}, one of the most widespread ST corpora and then we scale to the high-resource data condition allowed by the task organizers.
On MuST-C, we show that with the aid of a Connectionist Temporal Classification (CTC) auxiliary loss \cite{Graves2006ConnectionistTC} and compression \cite{gaido-etal-2021-ctc} in the encoder, our Conformer-based \cite{gulati20_interspeech} model 
can outperform -- to the best of our knowledge -- the previous best reported value of 25.3 BLEU by
\citet{inaguma-etal-2021-source},
even 
avoiding any 
additional pre-training or transfer learning.
Moreover, with the addition of a simple data filtering method, we achieve the new state-of-the-art score of 26.7 BLEU for a direct ST model that does not exploit external (audio or textual) resources.
Scaling to high-resource data conditions, we notice that the gap between an ASR pre-trained system and a system trained from scratch is closed only after a fine-tuning on
in-domain data.
Our submission to the offline task consists of an ensemble of
three models
that scores 32.2 BLEU on MuST-C v2 and 27.6 on IWSLT tst2020.

In the same vein of reducing
the overall training computational costs, we participated also in the simultaneous task using our best offline model and without performing any additional training do adapt it to the simultaneous scenario \citep{offline-simultaneous}.
The simultaneous version of our offline-trained model is realized by applying the wait-k strategy \citep{ma-etal-2019-stacl} with adaptive word detection from the audio input \citep{ren-etal-2020-simulspeech}
that determines
the number of words in a speech segment 
using the greedy prediction of the CTC.
Our SimulST model achieves competitive results on the MuST-C v2 test set compared to the last year systems, scoring 
25 BLEU at medium latency ($<2s$) and 30 BLEU at high latency ($<4s$) while 
keeping
low ($300-400ms$) the computation overhead and requiring no dedicated training.

\section{Competitive ST without Pre-training}

Before training systems on huge corpora, we conduct preliminary experiments on the MuST-C benchmark to find 
a 
promising setting
aimed at reducing the high computational costs of ST.
First, we validate on different architectures the 
finding of previous works \cite{gaido-etal-2021-ctc,papi-etal-2021-speechformer} that ST models trained with an additional CTC loss do not need an initialization of the encoder with that of an ASR model.
To this aim, 
we add a CTC loss \cite{gaido-etal-2021-ctc} whose targets are the lowercase transcripts without punctuation.\footnote{\mg{We add the CTC loss in the 8th encoder layer since \citep{gaido-etal-2021-ctc,papi-etal-2021-dealing} has demonstrated that it compares favourably with adding the CTC on top of the encoder outputs or in other layers \citep{bahar2019comparative}.}}
Second, we explore data selection mechanisms to increase model quality and reduce training time.
We always use the same hyper-parameters used in our final trainings for all systems (see Section \ref{sec:expset}) unless otherwise specified.

\subsection{Model Selection}

As a first step, we compare different architectures proposed for ST: ST-adapted Transformer \cite{wang2020fairseqs2t}, Conformer \cite{gulati20_interspeech}, and Speechformer \cite{papi-etal-2021-speechformer}.
In addition, we also test a composite architecture made
of a first stack of 8 Speechformer layers and a second stack of 4 Conformer layers. Hereinafter, we refer to this architecture as Speechformer Hybrid.
As a side note, we also experimented with replacing the ReLU activation functions in the decoder of our Conformer model with the squared ReLU,
in light of the recent findings on language models \citep{Primer} showing accelerated model convergence, decreased training time, and improved performance.
Unfortunately, these benefits were not observed in our experiments, as the introduction of the squared ReLU caused a small performance drop (-0.2 BLEU) and did not improve the convergence speed of the model. So, we do not consider this change in the rest of the paper.

In all
the
architectures, the encoder starts with two 1D convolutions. These layers compress the input sequence by a factor of 4 except for the Speechformer, where they do not perform any downsampling.
Indeed, the Speechformer relies on a modified self-attention mechanism (ConvAttention) with reduced memory requirements and shrinks the length of the input sequence only on top of 8 ConvAttention layers by means of the CTC-compression \cite{gaido-etal-2021-ctc} mechanism before feeding the sequence to 4 Transformer layers.
However, in a randomly initialized state, the CTC compression may actually not reduce the input sequence (or only slightly),
leading to OOM errors caused by the quadratic memory complexity with respect to the sequence length of the Transformer layers. For this reason \citet{papi-etal-2021-speechformer} initialize their encoder layers up to the CTC-compression module with a pre-trained model.
Since we aim at reducing the computational cost avoiding any pre-training, we introduce two methods that ensure a minimal compression factor of the input sequence after the CTC-compression:

\begin{itemize}
\item \textbf{Max Output Length}: if the sequence produced by the CTC compression is longer than a threshold (a hyper-parameter that we set to 1/4 of the maximum input sequence length\footnote{This ensures that the resulting sequences are not longer than the maximum length obtained by the Transformer and Conformer architectures after the two 1D convolutions.}), we merge \mg{(averaging them)} an equal number of consecutive vectors so that the final length of the sequence is inferior of the defined threshold. \mg{For instance, if the maximum input sequence length is 4,000, we set the threshold to 1,000; in this case, if a sample results in a sequence of length 2,346 after the CTC compression, we merge the first 3 vectors, then the vectors from the 4th to the 6th, and so on. We use 3 because it is the minimum compression factor that satisfies the length requirement.\footnote{\mg{A compression factor 2 would result in a sequence of length 1,173 -- higher than the 1,000 threshold -- while 3 produces a sequence of length 782.}}}
\item \textbf{Fixed compression}: for a given number of epochs $n_{E}$ 
(a hyperparameter)
the CTC compression is disabled and replaced by a fixed compression that averages 4 consecutive vectors. \mg{In this way, we directly control the length of the sequence after the compression, resembling the fixed compression performed by the initial 1D convolutional layers of Transformer and Conformer ST models.}
\end{itemize}

We choose the $n_{E}$ parameter of the fixed compression method among the values 6, 8, 10, and 12 according to the BLEU score\footnote{BLEU+case.mixed+smooth.exp+tok.13a+version.1.5.1} on the dev set. The best score was achieved with $n_{E}=10$ (24.16 BLEU), which was lower than the score obtained by the Max Output Length method (24.26 BLEU).
As such, in Table \ref{tab:model_sel} (\textit{w/o pretrain} column) we 
report the results of Speechformer and Speechformer Hybrid with the 
Max Output Length method.

The results show that the Speechformer-based models do need 
pre-training to reach their best 
scores while
Conformer and Transformer models achieve
comparable
translation quality 
avoiding the 
pre-training.
Specifically, the Conformer architecture with CTC compression obtains the best score without pre-training (25.5 BLEU) and has a negligible gap from the best result with pre-training (25.7 of Speechformer Hybrid).
We can hence confirm the statement that ASR pre-training can be avoided at barely no translation quality cost, and hereinafter we use the Conformer with CTC compression
without pre-training 
unless noted otherwise.
It is worth mentioning that the introduction of the CTC compression in the Conformer encoder 
does not only increase translation quality; also, it  reduces the RAM requirements and speeds up both the inference and training phases. Indeed, as
the sequence length is significantly reduced in the last encoder layers and in the encoder-decoder attention, less computations are required and the mini-batch size -- 
the number of samples processed in parallel -- can be increased.
Overall, this leads to save $\sim35\%$ of the training and inference time.

\begin{table}[bt]
\small
\centering
\begin{tabular}{l|cc}
    \hline
    \textbf{Model} & \textbf{w pretrain} & \textbf{w/o pretrain} \\
     \hline
     Transformer & 23.6 & 23.6 \\
     Speechformer & 24.5 & 24.3 \\
     Conformer & 24.8 & 24.8 \\
     \hspace{2mm} + CTC compr. & 25.6 & \textbf{25.5} \\
     Speechformer Hybrid & \textbf{25.7} & 24.9 \\
    \hline
\end{tabular}
\caption{SacreBLEU on the \textit{tst-COMMON} set of MuST-C v1 en-de.}
\label{tab:model_sel}
\end{table}

\subsection{Data Filtering}
\label{subsec:data_filter}

Easy methods to improve the quality of ST systems -- and deep neural networks in general -- consist in providing them with \textit{more} data or \textit{better} data.
The first approach comes at the cost of longer training time and higher computational 
requirements. This
makes the second approach more appealing
and in line with the overall goal and spirit of 
this work.
We hence focus on 
the definition of an efficient filtering strategy that improves the quality of our training data (and consequently of our models)
without additional computational costs.

We start from the observation that ST models estimate the probability of an output text given an input audio $p(Y|X)$, and a good ST model assigns a low probability to erroneous samples, which are outliers of the $p(Y|X)$ distribution.
Although
training a ST model only to filter the training data would be extremely computationally expensive, 
we decided to adopt this method as an upper bound for comparison with easier
and
feasible 
strategies.
In particular, 
for each sample in the training set, we computed the negative log-likelihood\footnote{The negative log-likelihood is defined as $- log(p(Y|X))$.} (NLL) with a strong ST model trained on all the data available for the competition (see Section \ref{sec:data}) as a proxy of the probability of the sample.
A high NLL means that a sample is unlikely, while a NLL close to $0$ means that the sample has a very high probability. Based on this, we can filter all the samples above a threshold to remove the least probable ones.
To set the threshold, we draw an histogram on all the training sets (see Figure \ref{fig:loss_based_filtering} in the Appendix) that leads to the following considerations:
\textit{i)} each dataset has a different distribution, making it difficult to define a threshold valid for all of them, and \textit{ii)} MuST-C has the highest NLL, meaning that it is more complex to fit for the model.

\sara{Through the approach described above, we selected the data of MuST-C - the dataset we used in these preliminary experiments - with a NLL greater than $4.0$. Upon a manual inspection of a sample of these selected data (5-10\% of the total),}
we noticed that two main categories were present: \textit{i)} bad source/target text alignments\footnote{In the MuST-C corpus, the alignments between transcripts and translations of the training set are automatically produced, hence misalignments and textual differences can be present.} (e.g. two sentences in the target translation are paired with only one in the transcript or vice versa),
and \textit{ii)} free (non-literal) translations. 
Instead, no cases of bad audio-transcript alignments were found (this was only a non-exhaustive manual inspection though), meaning that this problem is likely less widespread and 
impactful
than the textual alignment errors in the corpus.

\sara{These considerations motivated us to search for a feasible strategy that filter out the bad source/target text alignments. We}
first considered a simple method that discards samples with too high or low ratio between the target translation length (in characters) and the duration of the source audio.\footnote{In practice, we compute the number of characters divided by the number of $10ms$ audio frames.}
The corresponding histogram on the training data can be found in Figure \ref{fig:charvsframe} in the Appendix.
Looking at the plots, it emerges that this ratio is strongly dataset-dependent,
likely due to the high variability in speaking rate for different domains and conditions, thus making it hard to set good thresholds. For this reason, also supported by the finding of the 
manual inspection on the good quality of audio-text alignments discussed above, we turn to examine the ratio between the target translation length and the \textit{source transcript length}.\footnote{We used normalized transcript without punctuation, so the length of the target translation is on average 1.2X that of the source transcript.} Figure \ref{fig:charratio} in the Appendix shows 
its
histogram: in this case, the behavior is consistent on all datasets, making it easy to determine good values for the minimum and maximum ratio to admit (we set them to 0.8 and 1.6).

\begin{table}[hbt]
\small
\centering
\begin{tabular}{l|c}
    \hline
    \textbf{Model} & \textbf{BLEU} \\
    \hline
    Cascade \cite{bahar-2021-tight-integrated} & 25.9 \\
    Tight Integrated Cascade \cite{bahar-2021-tight-integrated} & 26.5 \\
    \hline
    \multicolumn{2}{c}{\textit{Without external data}} \\
    \hline
    SATE \cite{xu-etal-2021-stacked} & 25.2 \\
    BiKD \cite{inaguma-etal-2021-source} & 25.3 \\
    \hline
    \multicolumn{2}{c}{\textit{With external data}} \\
    \hline
    JT-ST \cite{tang-etal-2021-improving} & 26.8 \\
    Chimera \cite{han-etal-2021-learning} & 26.3 \\
    \hline
    \multicolumn{2}{c}{\textit{This work}} \\
    \hline
     Conformer + CTC compr. & 25.5 \\
     \hspace{2mm} + char-ratio filter. & 26.7 \\
     \hspace{2mm} + NLL-based filter. & 26.9 \\
    \hline
\end{tabular}
\caption{SacreBLEU on the \textit{tst-COMMON} set of MuST-C v1 en-de. Chimera uses additional speech and WMT14 \cite{bojar-etal-2014-findings}, while JT-ST uses only WMT14 as external resource.}
\label{tab:model_filter}
\end{table}

In Table \ref{tab:model_filter} we report the results of our filtering method
and we compare it with the upper bound of the NLL-based filtering strategy 
as well as with
previous works both 
under
the same data condition and with 
additional
external data.
First, we can notice that our simple method based on the target/source character ratio leads to a 1.2 BLEU gain, and has a very small gap (0.2 BLEU) with respect to
the upper bound exploiting a strong ST model for filtering.
Second,
our score (26.7 BLEU) is significantly higher than those reported by previous direct ST works in the same data condition and is on par or even outperforms those of models trained with the addition of external resources.
Finally, we compare the results of our model with those of the best cascade models reported in the same data conditions \cite{bahar-2021-tight-integrated}: the tightly-integrated cascade is close to our model (-0.2 BLEU), but ours also
benefits from the data filtering technique we just discussed.

To sum up, we managed to define a training recipe that enables reaching state-of-the-art ST results on MuST-C en-de (26.7 BLEU) with a single training step and involves: \textit{i)} the Conformer architecture, \textit{ii)} an auxiliary CTC loss and CTC-compression in the 8th encoder layer, and \textit{iii)} a simple yet effective filtering strategy based on the ratio between source and target number of characters.
In the following section, we discuss the application of this procedure in high-resource data conditions.

\section{Audio Segmentation Strategy}
\label{sec:audiosegm}
ST models are usually trained and evaluated in the ideal and unrealistic condition of audio utterances split at sentence level. As such, when fed with an unsegmented audio stream, they suffer from the mismatch between the training and inference data, which often results in significant performance drops. Accordingly, our last year submission \citep{papi-etal-2021-dealing} focused on reducing the impact of this distributional shift,
both by increasing the robustness of the model with a fine-tuning on a random re-segmentation of the MuST-C training set \cite{gaido2020contextualized}, and by
means of
a hybrid method for audio segmentation \citep{gaido-etal-2021-beyond}, which considers both the audio content and the desired length of the 
resulting
speech segments. The experiments showed that the two approaches accounted for complementary gains, both contributing to obtain our best scores.

Recently, \citet{tsiamas2022shas} presented a novel Supervised Hybrid Audio Segmentation (SHAS) with excellent results in limiting the translation quality 
drop. SHAS adopts a probabilistic version of the Divide-and-Conquer algorithm by \citet{potapczyk-przybysz-2020-srpols} that progressively splits the audio at the frame with highest probability of being a splitting point until all segments are below a specified length.
The probability of being a splitting point is estimated by a classifier fed with audio representations generated by wav2vec 2.0 \citep{BaevskiZMA20-wav2vec} and trained to approximate the manual segmentation of the existing corpora, i.e. to emit 1 for frames representing splitting points
and 0 otherwise. 
\mg{Since this approach involves a prediction with neural models of considerable size,} its superiority over the VAD-based ones comes with a significant computational cost and overhead.
In addition, SHAS is not
applicable to audio streams, as it requires the full audio to be available before start splitting. In the context of this competition, however, these limitations do not represent a significant issue.

\citet{tsiamas2022shas} compare SHAS with previous segmentation methods only using models trained on well-formed sentence-utterance pairs. In this work, we validate their findings also on models fine-tuned on randomly segmented data to check: \textit{i)} whether this fine-tuning brings benefits also with audio segmented with SHAS, and \textit{ii)} whether the gap between SHAS and other segmentation is closed or not by the fine-tuning.

\section{Simultaneous}
\label{sec:simultaneous}
In light of the recent work that questions the necessity of a dedicated training procedure for simultaneous model \citep{offline-simultaneous}, we participate in the Simultaneous task with the same model used for the Offline task.
Their finding is perfectly aligned with the spirit of this submission toward the reduction of training computational costs.
%
%
%
We determine when to start generating the output translation adopting
the wait-k strategy \citep{ma-etal-2019-stacl} that
%
%
simply prescribes to wait for $k$ words before starting to generate the translation, where $k$ is a hyper-parameter controlled by the user that can be increased or decreased to directly control the latency of the system.
The number of words in a given input speech is determined with an  adaptive word detection strategy \citep{ren-etal-2020-simulspeech}, because of its superiority over the fixed strategy \citep{ma-etal-2020-simulmt} in strong models trained in high-resource data conditions \cite{offline-simultaneous}.
Our adaptive word detection mechanism exploits the predicted output of CTC module in the encoder \citep{ren-etal-2020-simulspeech,zeng-etal-2021-realtrans} to count the number of words in the source speech.

 
The number of words to wait -- $k$ -- is not the only hyper-parameter that controls the wait-k strategy. Another important factor is \textit{how often} we check the number of uttered words that is the length of the \textit{speech segment}.
A short speech segment means that the system decides more frequently whether to wait for more input or to produce a part of output. This can reduce the latency, but it increases the number of forward passes through the encoder and hence the computational cost. In addition, a longer speech segment implies that the system takes decision with more context at disposal, possibly improving the quality.
For this reason,
we performed preliminary experiments exploring different speech segment dimensions (every $40ms$ ranging from $120ms$ to $720ms$) and we found $320ms$ and $640ms$ to be superior to other values.
Accordingly, we report the results of our systems for these two speech segment durations
varying the value of $k$ to achieve different latency.
In particular, we test our model with $k = \{1, 2, 3, 4, 5, 6, 7, 8, 9, 10, 11, 12, 13, 14\}$ in order to lie in the latency intervals prescribed by the Simultaneous Shared Task.\footnote{\url{https://iwslt.org/2022/simultaneous}}
The latency intervals are determined by the Average Lagging \citep{ma-etal-2020-simulmt} -- or AL -- on MuST-C v2 tst-COMMON and are: \emph{Low Latency} with $AL \leq 1000ms$, \emph{Medium Latency} with $AL \leq 2000ms$, and \emph{High Latency} with $AL \leq 4000ms$.
We use a standard AL-BLEU graph to report the system performance, where in the x axis we find the AL values ranging from $700ms$ to $4000ms$ and in the y axis the corresponding BLEU values.
Moreover, we also report the AL\textsubscript{CA}, the computational aware version of the AL metric \citep{ma-etal-2020-simulmt} accounting also for the computational time spent by the model during inference, in an AL\textsubscript{CA}-BLEU graph that will be used to additionally score the performance in the simultaneous task.

\section{Data}
\label{sec:data}

%
As training set, we use
the ASR and ST datasets allowed for the offline task,\footnote{\url{https://iwslt.org/2022/offline}} which are the same allowed for the simultaneous one.
The ASR data consist in 
\textit{(speech, transcript)} pairs that, in our case, are in English. The ST data consist in \textit{(speech, transcript, translation)} triplets from a source language (here English) to a target language (here German).
The ASR data we used are: LibriSpeech \citep{librispeech}, TEDLIUM version 3 \citep{tedlium}, Voxpopuli \citep{wang-etal-2021-voxpopuli}, and Mozilla Common Voice.\footnote{\url{https://commonvoice.mozilla.org/en/datasets}}
The ST data we used are: MuST-C version 2 \citep{Cattoni2020mustc-v2}, CoVoST version 2 \citep{wang-etal-2020-covost}, and Europarl-ST \citep{jairsan2020a}.

The ASR-native corpora were included in our ST training by applying
Sequence Knowledge Distillation \citep{kim-rush-2016-sequence,gaido-2020-on-knowledge} -- or SeqKD --, a popular data augmentation method used in the past IWSLT editions \citep{ansari-etal-2020-findings,anastasopoulos-etal-2021-findings} in which a teacher MT model is used to translate the source transcripts 
into the target language.
To avoid additional computational costs, we choose as MT teacher the freely available pre-trained model by \citet{tran2021facebook} for WMT2021 that was trained on the corresponding WMT2021 dataset \citep{akhbardeh-etal-2021-findings}, allowed by the IWSLT2022 Offline Task.
The SeqKD method was also applied to MuST-C v2 in order to augment the scarce ST 
available data.
As such, our training set comprised the synthetic data built using SeqKD and the native ST data, both filtered with the method described in Section \ref{subsec:data_filter}. The two types of data were distinguished by means of a tag pre-pended to the target text \citep{gaido-etal-2020-end,papi-etal-2021-dealing}.

\section{Experimental Settings}
\label{sec:expset}
All the models used for our participation were implemented
on Fairseq-ST \citep{wang2020fairseqs2t}.\footnote{Code available at: \url{https://github.com/hlt-mt/FBK-fairseq}.}
All the architectures (Transformer, Speechformer, Speechformer Hybrid, and Conformer) consist in 12 encoder layers and 6 decoder layers, 512 features for the attention layers and 2,048 hidden units in the feed-forward layers. 
We used 0.1 dropout for the feed-forward layer and attention layer. For Conformer convolutional layers we also apply 0.1 dropout and we set the kernel size to 31 for the point- and depth-wise convolutions.
We trained with the Adam optimizer \citep{DBLP:journals/corr/KingmaB14} ($\beta_1=0.9$, $\beta_2=0.98$). The learning rate was set to increase linearly from $0$ to $2e-3$ for the first 25,000 warm-up steps and then to decay with an inverse square root policy.
Differently, it was kept constant for
model fine-tuning, with a value of $1e-3$.
The vocabularies are built via SentencePiece models \citep{sennrich-etal-2016-neural}.
In our preliminary experiments only on MuST-C, the number of merge operations was set to 8,000 \cite{di-gangi-etal-2020-target} for the German translations and 5,000 \cite{wang2020fairseqs2t} for the lowercase punctuation-free English transcripts.
In the experiments on high-resource data condition, we doubled these values.
We normalize the audio features before passing them to our models with Cepstral Mean and Variance Normalization. Specifically, in offline ST the mean and variance are estimated at utterance level, while for simultaneous ST inference the normalization is based on the global mean and variance estimated on the MuST-C version 2 training set.

Trainings were performed on 4 NVIDIA A100 GPUs with 40GB RAM. We set the maximum number of tokens to 40k per mini-batch and 2 as update frequency for the Conformer with CTC-compression. The other models were trained with 20k tokens per mini-batch and 4 as update frequency.
We trained each model for
100,000 updates, corresponding to about 28 hours for the Conformer with CTC-compression. For offline generation, the maximum number of tokens was decreased to 25k, since we used a single K80 GPU with 12GB RAM and we applied the beam search strategy with \texttt{num\_beams=5}.
For simultaneous generation based on SimulEval \citep{ma-etal-2020-simuleval}, we used a K80 GPU and greedy search.

\section{Results}

In this section, we report our experiments in high-resource data conditions and we discuss our submission to the Offline (section $\ref{subsec:offline-res}$ and Simultaneous (section $\ref{subsec:simul-results}$) tasks.

\begin{table}[tb]
\small
\centering
\begin{tabular}{rl|c}
    \hline
    & \textbf{Model} & \textbf{BLEU} \\
     \hline
     I. & Conformer & 30.6 \\
     II. & \quad + in-domain fn & 31.6 \\
     III. & Conformer\_pretrain & 31.5 \\
     IV. & \quad + in-domain fn & \textbf{31.7} \\
     \hline
     V. & Ensemble (II, III) & 32.0 \\
     VI. & Ensemble (III, IV) & 31.7 \\
     VII. & Ensemble (II, IV) & \textbf{32.2} \\
    \hline
\end{tabular}
\caption{BLEU on MuST-C v2 tst-COMMON for Conformer with pretraining (\textit{Conformer\_pretrain}) and without it (\textit{Conformer}). We also report the scores after fine-tuning
on in-domain data (\textit{+ in-domain fn}).}
\label{tab:indomain}
\end{table}

\subsection{Offline}
\label{subsec:offline-res}

\begin{table*}[htb]
\centering
\small
\begin{tabular}{rl|cc|cc}
    \hline
    & \multirow{2}{*}{\textbf{Model}} & \multicolumn{2}{c|}{\textbf{Hybrid}} & \multicolumn{2}{c}{\textbf{SHAS}} \\
    \cline{3-6}
    & & tst-COMMON & iwslt2020 & tst-COMMON & iwslt2020 \\
     \hline
     1. & Conformer + in-domain fn & 27.4 & 23.8 & 30.3 & 26.4 \\
     2. & Conformer\_pretrain + in-domain fn & 28.1 & 24.4 & \textbf{30.4} & \textbf{26.8} \\
     \hline
     \multicolumn{6}{c}{\textit{with fine-tuning on resegmented data}} \\
     \hline
     3. & Conformer + resegm. fn & 28.3 & 25.2 & 29.3 & 26.1 \\
     4. & Conformer + in-domain fn + resegm. fn & 29.1 & 25.0 & 29.9 & 26.2 \\
     5. & Conformer\_pretrain + resegm. fn & 29.0 & 25.9 & 29.8 & 26.7 \\
     6. & Conformer\_pretrain + in-domain fn + resegm. fn & 29.0 & 25.7 & 29.7 & \textbf{26.8} \\
     \hline
     \multicolumn{6}{c}{\textit{Ensembles}} \\
     \hline
     7. & Ensemble (1, 2) & 28.6 & 24.7 & 30.9 & 27.2 \\
     8. & Ensemble (4, 6) & 29.7 & 26.0 & 30.5 & 27.2 \\
     9. & Ensemble (2, 6) & 28.9 & 25.7 & 30.8 & 27.4 \\
     10. & Ensemble (1, 2, 6) & 28.9 & 25.8 & \textbf{31.3} & \textbf{27.6} \\
    \hline
\end{tabular}
\caption{
BLEU scores of Hybrid and SHAS audio segmentation methods of the models with and without fine-tuning on re-segmented data (\textit{resegm. fn}) on the MuST-C v2 tst-COMMON and the IWSLT2020 test set.}
\label{tab:resegm}
\end{table*}

\paragraph{Fine-tuning on in-domain data.}  
In addition to training our models in the high-resource data condition, we also investigate whether fine-tuning on in-domain data brings advantages or not.
The results are reported in Table \ref{tab:indomain}. 
As we can notice, the Conformer with pre-training outperforms its version trained from scratch by 0.9 BLEU. However, when both the systems are fine-tuned on the in-domain data (rows II and IV), this difference becomes negligible (0.1 BLEU) meaning that the pre-training phase can be skipped in favor of a single fine-tuning step.
This might also suggest that the learning rate scheduler and the 
hyper-parameters we used -- tuned on
MuST-C corpus -- may be sub-optimal when a large amount of data is available. 
For time reasons, we did not investigate this aspect, which we leave to future work.
In addition, we compared several model ensembles: the Conformer with fine-tuning (II) and the pre-trained Conformer (III); the pre-trained Conformer (III) and the pre-trained Conformer with fine-tuning (IV); the Conformer with fine-tuning (II) and the pre-trained Conformer with fine-tuning (IV).
Our results show that ensembling the 
pre-trained Conformer and its fine-tuned version 
(VI) does not bring benefits, while selecting 
the Conformer without pre-training fine-tuned on in-domain data and the Conformer with pre-training
(V) leads to some improvements, which are 
enhanced when 
the two fine-tuned models 
are used (VII).
We also tested ensembles with more than 2 models without obtaining any advantage in terms of translation quality.

\paragraph{Fine-tuning on re-segmented data.}
As introduced in Section \ref{sec:audiosegm}, we tested
two audio segmentation methods: the \emph{Hybrid} segmentation 
\citep{gaido-etal-2021-beyond}, and the \emph{SHAS} segmentation 
\cite{tsiamas2022shas}. Also, we fine-tuned our \mg{ST} models on automatically re-segmented data to reduce the mismatch between train and evaluation conditions. The results are shown in Table \ref{tab:resegm}.
%
First, we notice that the SHAS segmentation method improves over the Hybrid one, with gains from 0.7 to 3.4 BLEU.
Secondly, we see that the fine-tuning on re-segmented data -- useful with the Hybrid segmentation -- becomes useless if using SHAS.
In fact, the best overall results are obtained using SHAS on a model that is not fine-tuned on resegmented data (row 2),
which scores 30.4 BLEU on the MuST-C v2 tst-COMMON and 26.8 BLEU on the IWSLT 2020 test set.
As such, we can conclude that fine-tuning on resegmented data is not needed if the audio is segmented with SHAS.

\paragraph{Ensembles.}
Since in the experiments on in-domain fine-tuning the best overall score was obtained by an ensemble of models, we compared the best combination (Ensemble VII in Table \ref{tab:indomain}) with other ensembles obtained by combining models fine-tuned on re-segmented data and models without this fine-tuning.
As we can see from rows 7-10 of Table \ref{tab:resegm}, the best scores are realized by adding a model fine-tuned on re-segmented data (6) to Ensemble VII, although the gap between all the ensembles is small on both test sets ($\leq 0.4$ BLEU).
This 3-models ensemble (10) obtained the best overall BLEU of 31.3 on MuST-C v2 tst-COMMON and 27.6 on IWSLT 2020 test set, outperforming by 1.6 BLEU the best result reported last year \citep{inaguma-etal-2021-source}.

\begin{table*}[htb]
\centering
\small
\begin{tabular}{ll|ccc|ccc}
    \hline
    & \multirow{2}{*}{\textbf{Model}} & \multicolumn{3}{c|}{\textbf{tst2022}} & \multicolumn{3}{c}{\textbf{tst2021}} \\
    \cline{3-8}
    & & ref2 & ref1 & both & ref2 & ref1 & both \\
     \hline
     Best direct IWSLT 2021 & \cite{bahar-etal-2021-without} & - & - & - & 22.6 & 18.3 & 31.0 \\
     Best cascade IWSLT 2021 & HW-TSC \cite{iwslt_2021} & - & - & - & 24.6 & 20.3 & 34.0 \\
     \hline
     \multicolumn{8}{c}{\textit{This work}} \\
     \hline
     primary & Ensemble (1, 2, 6) & \textbf{23.6} & \textbf{21.0} & \textbf{32.9} & \textbf{25.5} & \textbf{21.3} & \textbf{35.6} \\
     \hline
     contrastive1 & Ensemble (1, 2) & 23.4 & 20.6 & 32.5 & 25.4 & 20.9 & 35.2 \\
     contrastive2 & Conformer + in-domain fn & 22.8 & 20.1 & 31.6 & 24.5 & 20.2 & 33.9 \\
     \hline
\end{tabular}
\caption{
\mg{BLEU scores on the official blind tst2022 and tst2021 sets of our primary and contrastive submissions.}}
\label{tab:official_results}
\end{table*}

\paragraph{Offline Submissions.} 
Given the results of the Ensemble (1, 2, 6), we chose its output as our \emph{primary} submission for the Offline Shared task.
On the basis of the small performance drop on both test sets (0.4 BLEU) and 
to verify the possibility of avoiding the fine-tuning on re-segmented data,
we choose the Ensemble (1, 2) as 
\mg{\textit{contrastive}} submission. 
Lastly, we can notice that
the single Conformer model without pre-training 
(1)
falls behind the best Ensemble by only 1 BLEU 
for MuST-C v2 tst-COMMON and 1.2 BLEU
for IWSLT 2020 test set. 
This 
suggests
that 
users can be served with sound and competitive translations
even with a single model obtained with less than 30 hours of total training time on 4 GPUs.
To test this hypothesis, we sent the translations generated by the latter system as additional 
\mg{\textit{contrastive}} submission.
\mg{We report in Table \ref{tab:official_results} the official results for the tst2022 and tst2021 sets. The scores confirm our findings that the gap between the best ensemble and the single model without pre-training is limited to less than 1 BLEU. Most significantly, this single model outperforms the best direct system reported last year \cite{bahar-etal-2021-without} by 1.9 BLEU on the two single references and 2.9 BLEU on both references. Our primary submission increases these gains to 2.9-3.0 BLEU on the single references and 4.6 BLEU on both references, and beats the best cascade system from last year campaign (HW-TSC -- \citealt{iwslt_2021}) by 0.9-1.0 BLEU on the single references and 1.6 BLEU on both references.}
All in all, we can conclude that this work has shown that a lightweight training procedure
is possible without dramatically sacrificing the quality and competitiveness of the system. We believe that our results are promising for future works in this direction.

\begin{figure}[tb]
     \centering
     \begin{subfigure}[b]{0.475\textwidth}
         \centering
         \includegraphics[width=\textwidth]{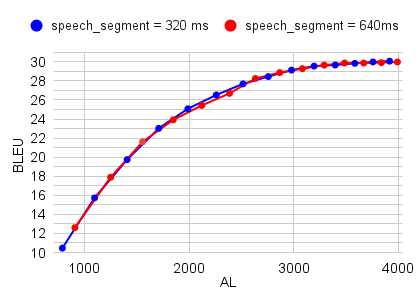}
     \end{subfigure}
     \hfill
     \begin{subfigure}[b]{0.475\textwidth}
         \centering
         \includegraphics[width=\textwidth]{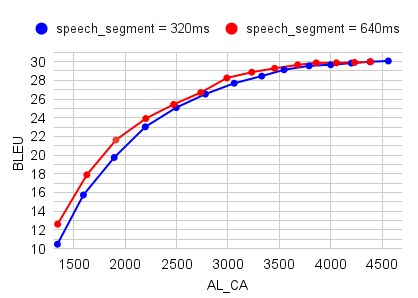}
     \end{subfigure}
        \caption{AL- and AL\textsubscript{CA}-BLEU curves on MuST-C v2 tst-COMMON.}
        \label{fig:simultaneous}
\end{figure}

\subsection{Simultaneous}
\label{subsec:simul-results}
For the SimulST task participation we 
use the 
best performing offline
model, namely the Conformer with pre-training and fine-tuning on in-domain data, to which we 
apply the wait-k policy with adaptive word detection.
The AL- and AL\textsubscript{CA}-BLEU graphs are shown in Figure \ref{fig:simultaneous}.

As we can see from the AL-BLEU graph, the systems with speech segment $320ms$ and $640ms$ have similar behaviour in terms of quality.
The main difference between them is 
the minimum latency from which they start:
the system with speech segment $320ms$ starts 
at an AL of about $800ms$ while the system with speech segment $640ms$ starts 
at about $900ms$. On average, if the $k$ value increases, the AL increases 
by $300ms$ for both systems, with a wider latency interval at the beginning that
progressively 
shrinks at high latency values.
In spite of this, the system with speech segment $320ms$
achieves
the highest BLEU slightly before the \emph{Medium Latency} (25.1) and \emph{High Latency} thresholds (30.1), making it the best candidate for 
submission.
If we look at the AL\textsubscript{CA}-BLEU graph, the results partially change because the system with speech segment $640ms$ 
has a
lower computational burden, achieving up to 2 BLEU points improvement at low latency against the other system. Thus, looking at the computational aware metric, the best candidate is the system with speech segment $640ms$. We can conclude that $320ms$ is the best speech segment value for the AL ranking while $640ms$ is the best for the AL computational aware version.
Since the organizers encourage multiple submissions, we participate with both the speech segment values.

\section{Conclusions}
We described the FBK participation in the IWSLT 2022 offline and simultaneous tasks \citep{iwslt:2022}. 
Our focus was to build a system with the least number of training steps but capable of obtaining competitive results with state-of-the-art models, which typically undergo
complex and longer training procedures.
To this aim, we \emph{i)} showed that ASR pre-training of the encoder can be avoided without a significant
impact on the final system performance, \emph{ii)} 
proposed a simple yet effective data filtering technique to enhance translation quality while reducing the training time, and \emph{iii)} compared different solutions to deal with automatic audio segmentation at inference time.
Our results on the 
IWSLT2020 test set indicate that
a single Conformer-based model without pre-training
falls behind our best model ensemble by only 1.2 BLEU and 
outperforms the best score reported last year
by 0.4 BLEU.
\mg{The same trend occurs on the blind tst2021 and tst2022 sets, with a 0.8-1.1 BLEU gap from our best model ensemble, which in turn beats by $\sim$1 BLEU the best reported result last year.}
These promising results are also confirmed in the simultaneous scenario in which, using the offline-trained model without any adaptation 
for the simultaneous task, we reach good quality-latency balancing, especially in the more realistic computational aware evaluation setting.

\section{Acknowledgements}
This work has been supported by the ISCRA-B project \textit{DireSTI} granted by CINECA, and by Amazon Web Services.
\mg{The submission to the simultaneous track has been carried out as part of the project Smarter Interpreting (\url{https://kunveno.digital/}) financed by CDTI Neotec funds.}

\bibliography{anthology,custom}
\bibliographystyle{acl_natbib}

\appendix

\begin{figure*}[tb]
     \centering
     \includegraphics[width=0.8\textwidth]{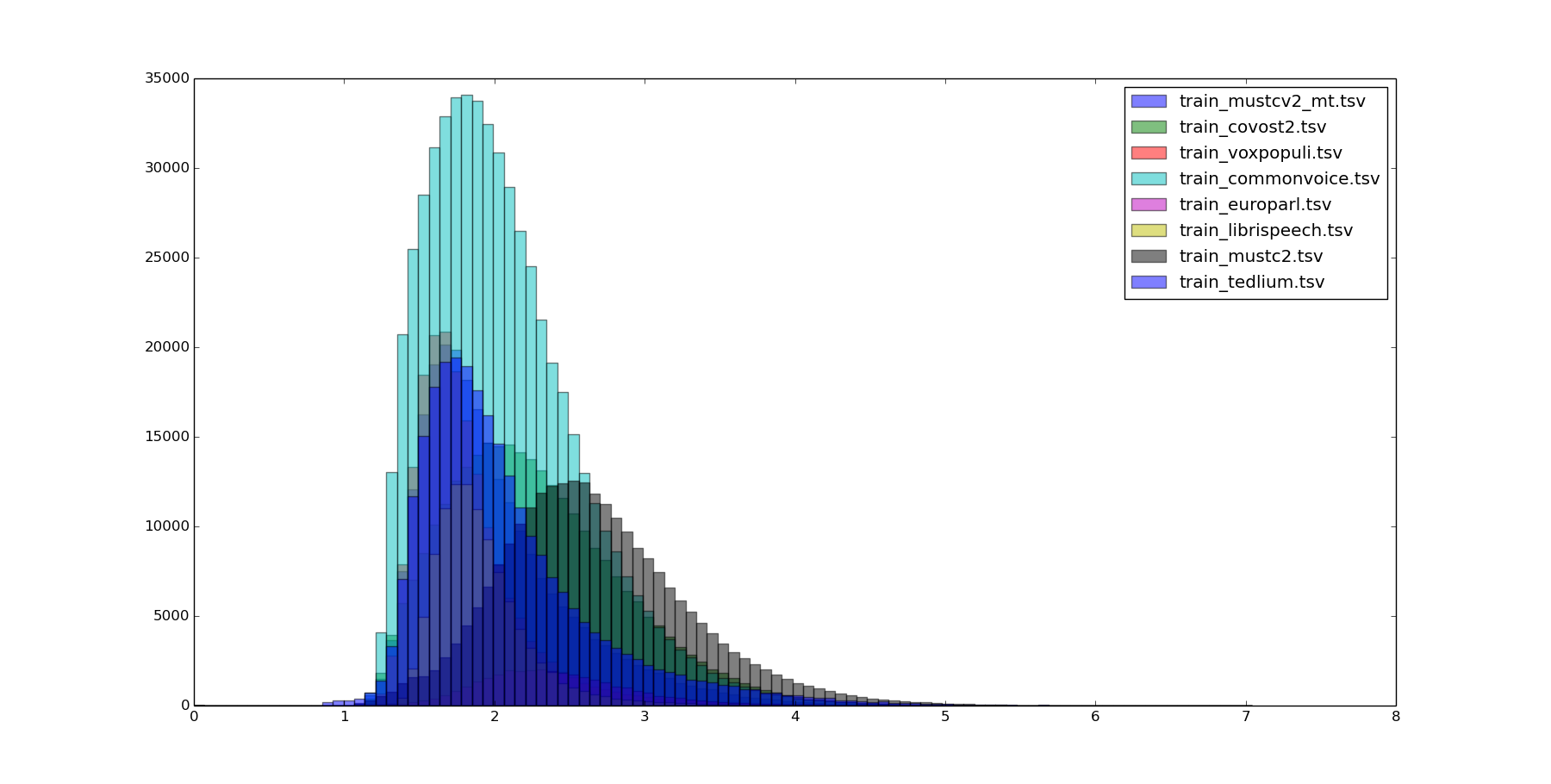}
    \caption{\label{fig:loss_based_filtering}Histogram of the negative log-likelihood (NLL) of the samples for all the training set of the competition. The ST model used to estimate the NLL has been trained on all the data and was scoring 29.6 BLEU on MuST-C.}
\end{figure*}
\begin{figure*}[tb]
     \centering
     \includegraphics[width=0.8\textwidth]{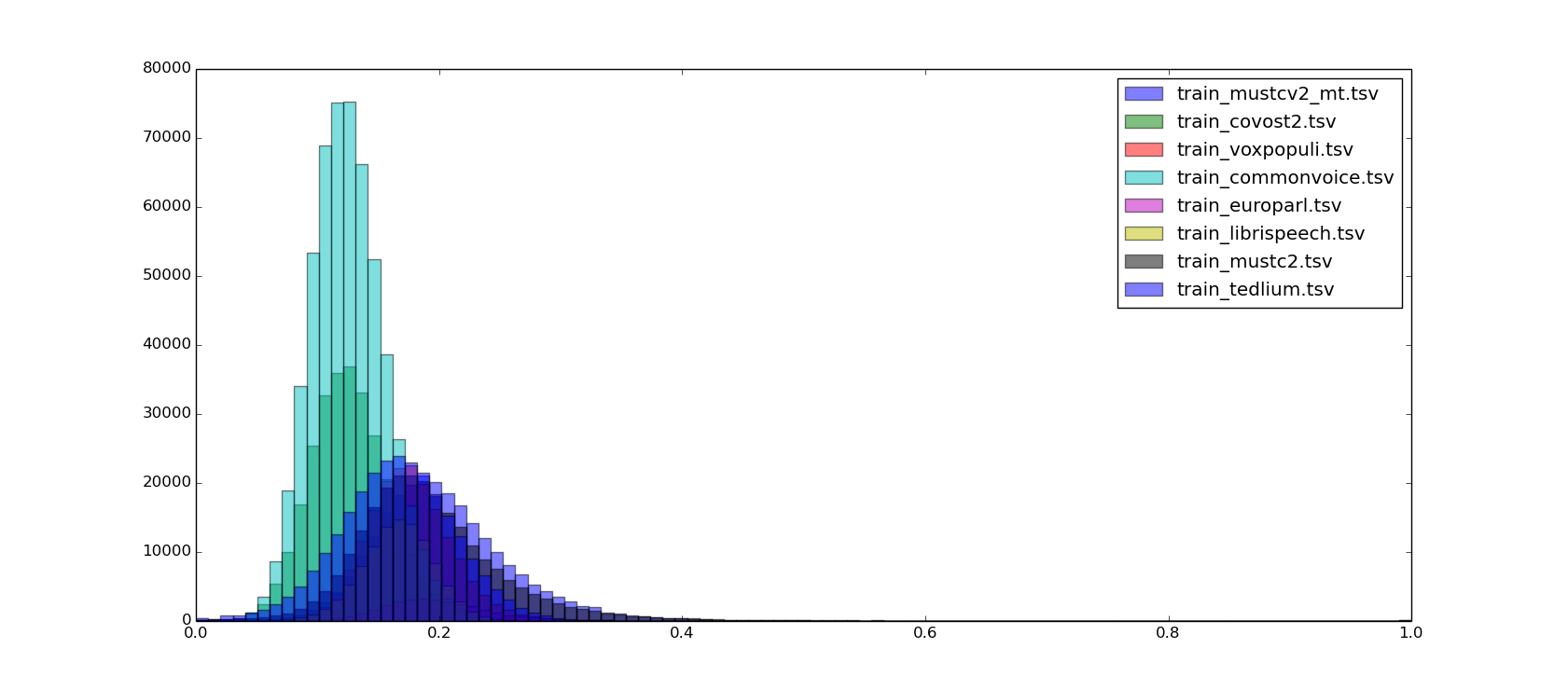}
    \caption{\label{fig:charvsframe}Histogram of the ratio between the number of target translation character and 10ms audio frames for all the training set of the competition.}
\end{figure*}
\begin{figure*}[tb]
     \centering
     \includegraphics[width=0.8\textwidth]{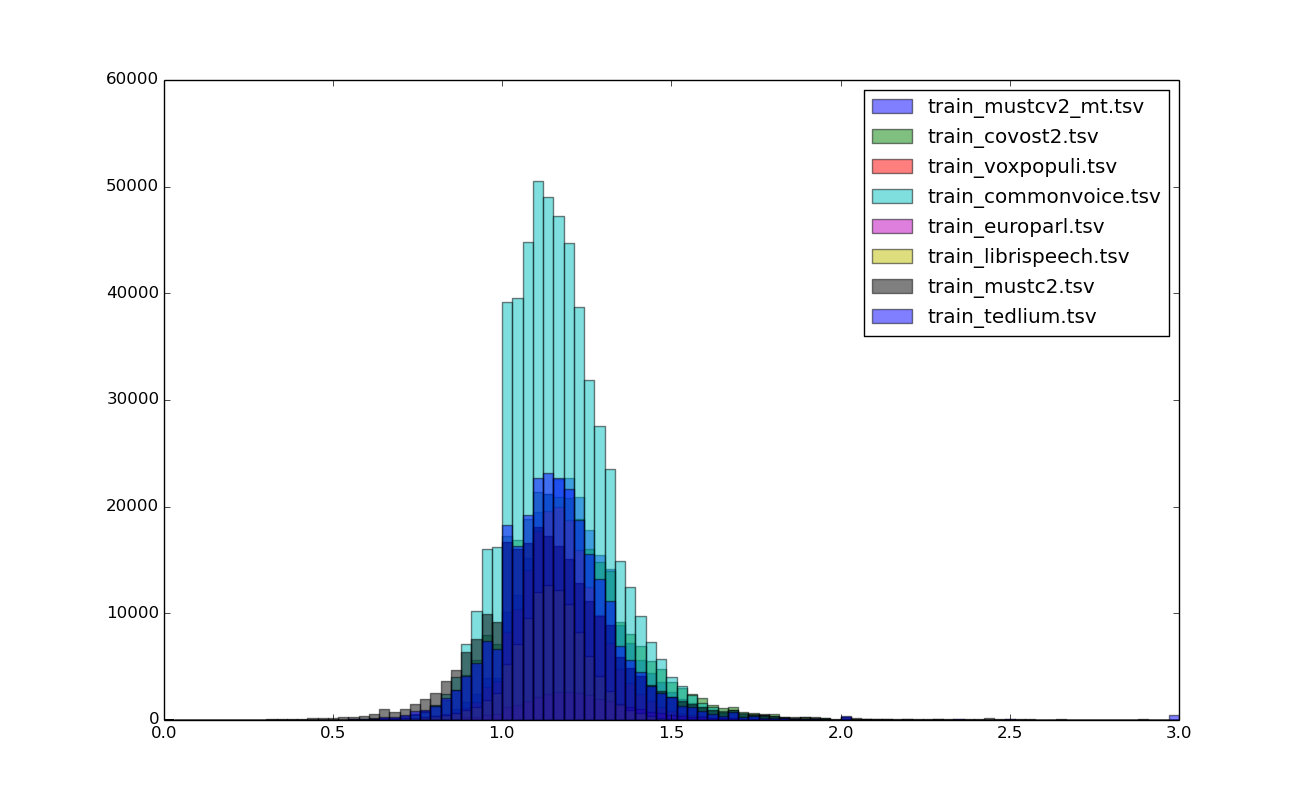}
    \caption{\label{fig:charratio}Histogram of the ratio between the number of characters in the target translation and the source punctuation-free transcript for all the training set of the competition.}
\end{figure*}

\section{Dataset Statisctics for Data Filtering}
\label{sec:appendix_df}

In this Section we report the histograms created when defining our data filtering mechanism (Section \ref{subsec:data_filter}).


\end{document}